\documentclass[conference]{IEEEtran}
\usepackage{cite}

\usepackage[normalem]{ulem}
\useunder{\uline}{\ul}{}

\usepackage{multirow}

\ifCLASSINFOpdf
  \usepackage[pdftex]{graphicx}
\else
\fi
\usepackage[shortlabels]{enumitem}

\usepackage{amsmath}
\usepackage{algorithmic}

\usepackage{array}
\usepackage{multirow}
\usepackage[table,xcdraw]{xcolor}

\ifCLASSOPTIONcompsoc
 \usepackage[caption=false,font=normalsize,labelfont=sf,textfont=sf]{subfig}
\else
 \usepackage[caption=false,font=footnotesize]{subfig}
\fi
\usepackage{url}

\usepackage{amsfonts}
\usepackage{multirow}
\usepackage{hyperref}
\usepackage{multirow}
\usepackage[table,xcdraw]{xcolor}
\usepackage[table,xcdraw]{xcolor}
\usepackage{subfig}

\begin{document}
%
\title{Leveraging Log Instructions in Log-based Anomaly Detection}


\author{\IEEEauthorblockN{Jasmin Bogatinovski\IEEEauthorrefmark{1},
		Gjorgji	Madjarov\IEEEauthorrefmark{3},
		Sasho Nedelkoski\IEEEauthorrefmark{1},
		Jorge Cardoso\IEEEauthorrefmark{2} and
		Odej Kao\IEEEauthorrefmark{1}}
	\IEEEauthorblockA{\IEEEauthorrefmark{1} Technical University Berlin, Berlin, Germany, Email: {jasmin.bogatinovski@tu-berlin.de}}
	\IEEEauthorblockA{\IEEEauthorrefmark{2} Huawei Munich Research, Munich, Germany}
	\IEEEauthorblockA{\IEEEauthorrefmark{3}  University Ss Cyril and Methodius,
		Skopje, North Macedonia}
}

\maketitle

\begin{abstract}
Artificial Intelligence for IT Operations (AIOps) describes the process of maintaining and operating large IT systems using diverse AI-enabled methods and tools for, e.g., anomaly detection and root cause analysis, to support the remediation, optimization, and automatic initiation of self-stabilizing IT activities. The core step of any AIOps workflow is anomaly detection, typically performed on high-volume heterogeneous data such as log messages (logs), metrics (e.g., CPU utilization), and distributed traces. In this paper, we propose a method for reliable and practical anomaly detection from system logs. It overcomes the common disadvantage of related works, i.e., the need for a large amount of manually labeled training data, by building an anomaly detection model with log instructions from the source code of 1000+ GitHub projects. The instructions from diverse systems contain rich and heterogenous information about many different normal and abnormal IT events and serve as a foundation for anomaly detection. The proposed method, named ADLILog, combines the log instructions and the data from the system of interest (target system) to learn a deep neural network model through a two-phase learning procedure. The experimental results show that ADLILog outperforms the related approaches by up to 60\% on the F$_1$ score while satisfying core non-functional requirements for industrial deployments such as unsupervised design, efficient model updates, and small model sizes. 
\end{abstract}



\IEEEpeerreviewmaketitle
\section{Introduction}
IT infrastructures in numerous application fields consist of thousands of networked software (microservices) and hardware (e.g., IoT, Edge) components. The uninterrupted and correct interaction is crucial for the functionality of the overall system and the deployed applications. However, this IT complexity combined with the required QoS guarantees (e.g. maximal latency) increasingly overwhelms the IT operators in charge. The current trends of agile software development with hundreds of updates and daily deployments further exacerbate the operational challenges. The holistic overview, operation, and maintenance of the IT infrastructure grow even more challenging when additionally it is affected by unforeseen factors such as failures, software errors, security breaches, or external environmental events. Companies react to these threats by employing additional site reliability engineers (SREs) as well as by deploying AI-enabled methods for IT operations (AIOps)~\cite{Notoro2021}. 

The AIOps methods collect and analyse plenty of IT system information -- metric data (e.g., CPU utilization), logs, and traces (paths of function calls) to detect anomalies, locate their root causes and remediate them. The diverse AIOps techniques enable fast, efficient and effective prevention of upcoming failures, aiming to minimize their hazardous effects during the daily operational activities~\cite{Notoro2021}. 

In this paper, we focus on anomaly detection in the context of AIOps, as a core step towards enhancing fault tolerance: the earlier an anomaly is detected, the more time is available to prevent the failure and mitigate the impact on the QoS. We focus on system log messages (logs) as semantically rich data written by humans for humans. The logs allow a more insightful analysis and interpretation than, e.g., metric data~\cite{LogClass2021}. For example, a sharp increase in the \textit{network packet loss} (a commonly used metric for network monitoring) only indicates a problem with the network, but it does not provide a clue why it happens. In comparison, logs give semantically meaningful clues for the anomaly. For example, when a switch generates the log “System is rebooting now.”, the operator detects that the switch is failing (potentially anomalous) and obtain a clue that the potential anomaly is caused by switch rebooting.


Logs are generated from log instructions that developers insert in the source code (e.g., log.info(”VM took \%f seconds to spawn.”,~createSeconds)) to visualise important system events and to create hints for the operators that run the system as a black-box~\cite{Hassan2020}. The log instructions are commonly composed of static text (log template), variable parameters of the event (e.g., createSeconds), and log level giving information about the severity level of the event (e.g., "info", "fatal", "error"). The log levels come at different granularity, conditioned on the used programming languages and logging libraries. The lower log levels such as "info" are usually used when describing \textit{normal} state or state transitions, e.g., "Successful connection.". In contrast, higher log levels such as "error", "critical", or "fatal" commonly accompany events that describe \textit{abnormal} states or state transitions, e.g., "Machine failure". Therefore, the log levels encode rich expert information for manual detection of anomalous events, frequently used in today's operational practices~\cite{Hassan2020}. For example, to diagnose an anomaly, operators commonly use manual search for logs with higher levels such as "error", "critical", or "fatal"~\cite{LogCluster}.




Owning to the ever-increasing IT system complexities, logs are constantly generated in large volumes (e.g., up to several TB per day~\cite{He2020SurveyLogMining}). The emergence of complexity makes the manual log-based anomaly detection time-consuming~\cite{Hassan2020}, prompting the need for automation~\cite{surveyLogAnalysis, surveyDL}. Thereby, automatic methods for log-based anomaly detection are increasingly researched and adopted~\cite{DeepLog, DT, PCA, LogRobust, CNN, HuaweiFLAP, SwissLog}. Current methods are commonly grouped into two families, i.e., supervised and unsupervised~\cite{surveyLogAnalysis}. Existing supervised methods depend on manually labeled training data. Due to the constant evolution of the software systems~\cite{LogRobust}, the supervised methods require a repetitive, \textbf{time-expensive labeling process}, which is oftentimes practically challenging and infeasible~\cite{He2020SurveyLogMining}. The unsupervised methods mitigate the labeling problem by modeling with logs from normal system states and detecting any significant deviations from the modeled normality state as anomalies. However, the lack of explicit information about anomalous logs during modeling leads to \textbf{limited input representation}, reducing their detection performance, and questioning their practical usability~\cite{surveyLogAnalysis}. 

To address the two challenges, we propose ADLILog. The central idea of the method is to use data from public code projects (e.g., GitHub) alongside the data from the system of interest (\textit{target system}) when learning the anomaly detection model. Since the public code projects contain numerous log instructions for diverse normal and abnormal events, we assume that they may encode rich anomaly-related information. Following the usage of the log levels for manual log anomaly detection, we considered grouping the instructions based on the log levels to extract anomaly-related information. Specifically, we created two severity level groups from the log instructions based on their log levels -- "normal" (composed of "info") and "abnormal" (composed of "error", "fatal", and "critical"). To verify our assumption, we conducted a study to examine the anomaly-related language properties between the two groups (i.e., diversity in the vocabulary and the sentiment of the words). The study results show that the two groups extract anomaly-related information that can be used as a basis for anomaly detection. Based on this observation, we introduce ADLILog, which uses the anomaly-related information alongside the target system data to learn a deep learning anomaly detection model through a two-phase learning procedure. 

An important advantage of ADLILog is that, by having access to "normal" and especially "abnormal" event descriptions from many different software systems, it learns a model by supervised learning objectives, without the need for target-system log labels (i.e., its unsupervised method). Thereby, ADLILog eliminates the need for \textbf{time-expensive} labeling while preserving the advantage of supervised modeling. The latter addresses the challenge of \textbf{limited input representation} during modeling. To prove the quality of detection, we extensively evaluate ADLILog against seven related methods on two widely used benchmark datasets and demonstrate that our method outperforms the supervised methods by 5-24\%, and the unsupervised by 40-63\% on F$_1$ score. The datasets\footnote{\href{https://zenodo.org/record/6376763}{https://zenodo.org/record/6376763}} and method implementation\footnote{\href{https://github.com/ADLILog/ADLILog}{https://github.com/ADLILog/ADLILog}} are available as open-source for fostering the research on this practically relevant problem.

The remaining of the paper is structured as follows. Section~\ref{sec3} presents our study that examines the potential of the log instructions to aid anomaly detection. Section~\ref{sec4} introduces ADLILog. Section~\ref{sec5} gives the experimental results. Section~\ref{rw} discusses the related work. Section~\ref{sec6} concludes the paper and gives directions for future work.

\section{Examining the potential of log instructions for log-based anomaly detection}~\label{sec3}

In this section, we examine the potential of the log instructions to aid anomaly detection. We start with our observation that there exist two log instructions severity groups, based on their log levels, i.e., "normal" ("info") and "abnormal" ("fatal", "critical", and "error"). Following the usages of the log levels for anomaly detection~\cite{LogCluster}, we assume that the static texts of the instructions have complementary properties concerning the two severity level groups, preserving anomaly-related information. To study the validity of the assumption, we analyze two language properties of the word combinations (n-grams) in the log instructions static texts with respect to the two groups. Specifically, by studying the n-gram uniqueness among the groups, we examine the differences in the vocabulary used to describe normal/abnormal events. By relating the n-grams with the expressed intent (e.g., positive intent relates to normal system state), we examine the semantic diversity between the groups, i.e., if the n-grams express positive (normal state transition) or negative (abnormal state transition) intents. In the following, we first describe the 1) \textit{log instruction collection procedure} and then present the 2) \textit{uniqueness} and the 3) \textit{sentiment} analyses of the log instructions static texts. 

\subsection{Log Instruction Collection and Processing}~\label{SLP} For the starting point of the analysis, we created a representative dataset by collecting log instructions from the source code of more than 1000 public code projects from~\href{http://github.com}{GitHub}. We included a wide spectrum of domains and programming languages (Python, Java, C++), covering different log instruction types. The heterogeneity enables us to examine the vocabulary diversity and semantic properties used in describing normal and abnormal events across systems. That way, we consider diverse logging styles and a wide range of events, with complementary severity levels. To account for the reliability in the log level assignment, we selected projects with more than a 100-stars and at least 20 contributors. The collection procedure resulted in more than 100.000 log instructions.  

Afterwards, we process the log instructions by extracting the log levels and the static texts to represent their severity levels and the event descriptions. The diverse programming languages use different names for the log levels. Therefore, as a first step, we unify all the log levels. We preprocess the static texts by applying several preprocessing techniques, similar to related works~\cite{28}, including lower-case word transformation, splitting the static texts on whitespace, removing placeholders, removing ASCII special characters and stopwords from the Spacy English dictionary~\cite{spacy}. We refer to this data as \textbf{Severity Level (SL)} data. It is a set of tuples from two elements -- (1) the static text of log instruction, and (2) the severity group based on the aforenamed log level to severity group mapping (e.g., ("machine error", "abnormal")). We used the SL data to conduct the log instruction examination study. Similar to related log instruction analysis studies~\cite{PinjiHe2018}, we extracted the n-grams from the static text by varying the value for the n parameter in the range $n=\{3, 4, 5\}$. An n-grams analysis shows that many n-grams appear once. To eliminate the impact of the rare n-grams on the analysis, we considered the n-grams that appear more than three times~\cite{PinjiHe2018}. 

\subsection{Log Instructions Static Texts Uniqueness Analysis} 
Intuitively, when describing abnormal events, the static text typically contains n-grams like "failure" or "error connection", as opposed to normal events, where n-grams like "successful" and "accepted" are more likely to appear. Therefore, we assume that the log instructions static texts of the two severity level groups share different, partially overlapping vocabularies. To verify this, we considered an approach from information theory that defines the amount of information uncertainty in a message~\cite{informationtheory}. In our case, we analyze the relation of the n-grams with the two severity groups. At first, given an n-gram (e.g., "machine failure"), there is high uncertainty for the assigned severity group. As we receive more information for the n-gram (e.g., new logs with the n-gram "machine failure"), its uncertainty concerning the associated severity group is reduced. For example, if the n-gram "machine failure" is associated five times with the "abnormal" and one time with the "normal" severity group, we have low uncertainty. In contrast, if another n-gram, e.g., "verifying connection" is associated three times with the "abnormal" and three times with the "normal" group, the n-gram uncertainty is high. To measure the uncertainty, we used Normalized Shanon's entropy~\cite{informationtheory}. We calculated the entropy for each n-gram and reported the key statistics of the n-grams entropy distribution.

\begin{table}[!h]
\centering
\caption{Log Instructions Static Texts Uniqueness Analysis Results}
\label{tab:ngramun}
\resizebox{\columnwidth}{!}{
\begin{tabular}{c|c|c|c|c|c}
\hline
                                                                      & Min  & 1st Qu. & Median & 3rd Qu. & Max  \\ \hline
\begin{tabular}[c]{@{}c@{}}Average Entropy\end{tabular} & 0.00 & 0.00    & 0.00   & 0.27    & 0.51 \\ \hline
\end{tabular}
}
\end{table}

\tablename~\ref{tab:ngramun} summarizes the key properties of the n-gram entropy distribution. It is seen that the median of the distribution is 0. This means that the majority of the n-grams are associated with only one of the two severity groups. Thereby, \textit{the two severity groups are characterized with a rather unique vocabulary.} While this analysis gives information about the uniqueness of the vocabularies, it does not account for the type of intent expressed with the n-grams. To investigate the expressed event intent, we made an n-gram sentiment analysis (where the sentiment is used to quantify the intent type, i.e., positive or negative), given in the following.

\subsection{Log Instructions Static Texts Sentiment Analysis} To evaluate the n-gram sentiment concerning the two severity groups, we considered a pretrained sentiment analysis model from Spacy~\cite{spacy}. We justify the applicability of the sentiment model by pointing to the observed similarities between general language and logs (as short texts)~\cite{PinjiHe2018}. Since the sentiment model is trained on diverse language texts, it has learned notions of positive, neutral or negative intent. We run the n-grams through the model to obtain the sentiment score. We used the sentiment score to categorize the n-grams into three categories, i.e., positive, negative and neutral. We relate the events from the "normal" severity group with positive intent because they describe a successful state or state transition. Similarly, we relate the "abnormal" group with a negative intent because it describes unsuccessful system state or state transition. The third category contains n-grams with neutral intent, i.e., events without strongly expressed intent.  

\begin{table*}[!h]
\centering
\caption{Log Instructions Static Texts Sentiment Analysis Results}
\label{tab:sentiment}
\resizebox{\textwidth}{!}{
\begin{tabular}{c|ccc|ccc|ccc}
\hline
Sentiment      & \multicolumn{3}{c|}{Positive}                                          & \multicolumn{3}{c|}{Negative}                                          & \multicolumn{3}{c}{Neutral}                                           \\ \hline
Severity Group & \multicolumn{1}{c|}{Normal}  & \multicolumn{1}{c|}{Abnormal} & Shared & \multicolumn{1}{c|}{Normal}  & \multicolumn{1}{c|}{Abnormal} & Shared & \multicolumn{1}{c|}{Normal}  & \multicolumn{1}{c|}{Abnormal} & Shared \\ \hline
N-gram Coverage [\%]               & \multicolumn{1}{c|}{66.94\%} & \multicolumn{1}{c|}{28.13\%}   & 4.93\% & \multicolumn{1}{c|}{23.13\%} & \multicolumn{1}{c|}{69.75\%}   & 7.12\% & \multicolumn{1}{c|}{46.98\%} & \multicolumn{1}{c|}{43.43\%}   & 9.59\% \\ \hline
\end{tabular}
}
\end{table*}

\tablename~\ref{tab:sentiment} summarize the results of the n-gram sentiment analysis. For each of the three sentiment categories, we show the percentages of the n-grams concerning the two severity groups. In the positive intent category 66.94\% of the n-grams are associated with the normal severity group, and 28.13\% are related to the abnormal severity group. In contrast, from the n-grams associated with negative intent, 69.75\% are associated with the abnormal group, 23.13\% are associated with the normal severity group, and 7.12\% are shared between the two. These two observations show that there exists a relationship between the normal group and positive intent, and the abnormal group and the negative intent. Therefore, the proposed severity log level grouping aligns with human intuition when expressing positive and negative sentiments. \textit{Structuring the static text of the log instructions by their log levels in the proposed way extracts anomaly-related information.} Combining this observation with the uniqueness in the vocabularies between the two severity groups demonstrates that SL data has rich anomaly-related properties, which can serve as a foundation for anomaly detection.

\section{ADLILog: Log-based Anomaly Detection by Log Instructions}~\label{sec4}
\begin{figure*}[!t]
\centering
\includegraphics[width=0.75\textwidth]{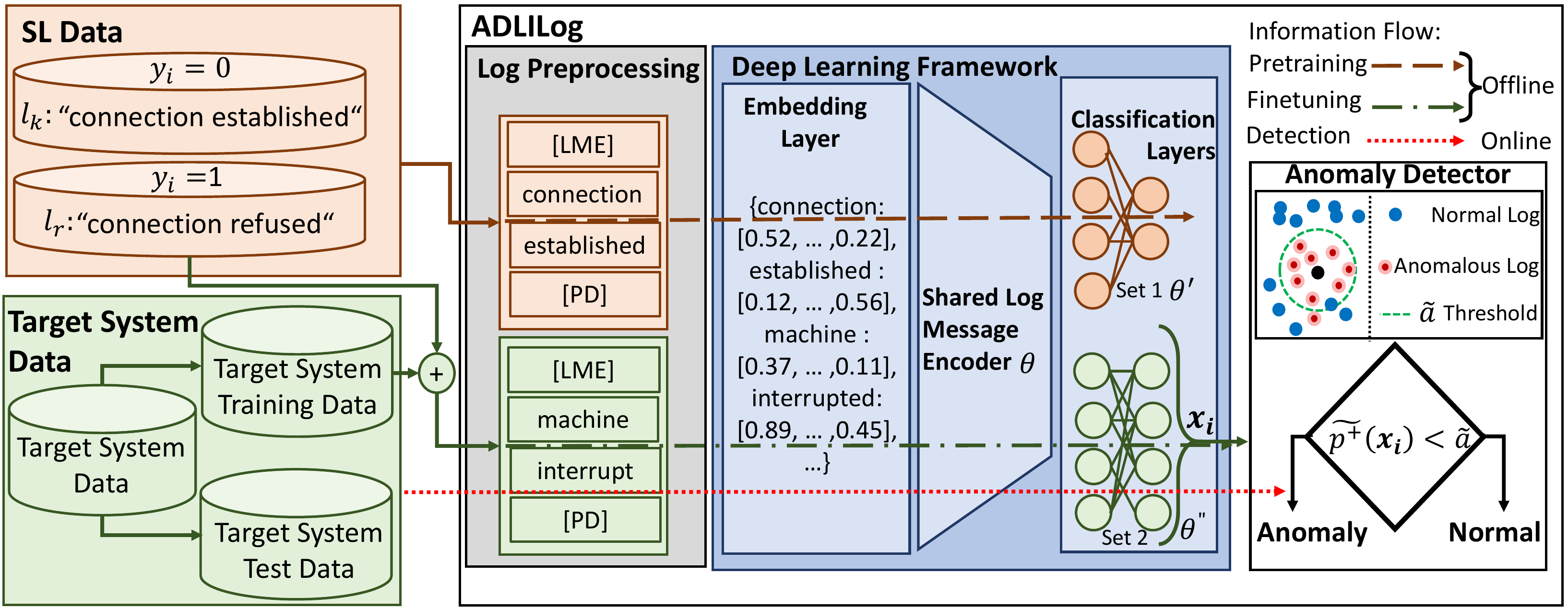}
\caption{ADLILog: Detailed design of the log anomaly detection method}
\label{fig:arch}
\end{figure*}

Following the affirmative observations about anomaly-related information encoded in the SL data from the examination study, in this section, we introduce ADLILog as an unsupervised log-based anomaly detection method. \figurename~\ref{fig:arch} illustrates the overview of the approach. Logically, it is composed of (1) \textit{log preprocessing}, (2) \textit{deep learning framework} and (3) \textit{anomaly detector}. The role of the \textit{log preprocessing} is to process the raw logs by carefully selecting preprocessing transformations that expose rich information for the deep learning framework. The \textit{deep learning framework}'s goal is to learn and output useful log representations for the target-system logs. It does so by training a deep neural network model with a sequential two-phase learning process (pretraining and finetuning), during which data from the target-system logs and the SL data are used. The \textit{anomaly detector} detects if the input target-system logs are normal or anomalous. In the following, we describe the three components of ADLILog.

\subsection{Log Preprocessing}~\label{implementation}
The raw target-system logs are characterized by high noise due to the parameter values generated during system runtime (e.g., IP address, endpoints, numerical parameters). The log noise can significantly affect the anomaly detection performance~\cite{surveyLogAnalysis}. Therefore, the log preprocessing aims to reduce the noise by applying a set of preprocessing steps. To that end, we start by removing all path endpoints (e.g., /home/spelce1/HPCCIBM/bin/) and split the static text using whitespaces into singleton items we call tokens. The tokens with numeric values most often denote variable parameters that are not relevant for the semantics of the logs. We consider them as noise and remove them. Similar to the preprocessing for the SL data, we apply Spacy and remove all ASCII special characters (e.g., $\$$), the stopwords (e.g., \texttt{is} and \texttt{the})~\cite{spacy} and transformed each character into a lower case, following related work~\cite{28}. Notably, as previously described, the SL data is already preprocessed by a similar set of operations making the preprocessing uniform. In addition, each log is prepended with a dedicated Log Message Embedding ([LME]) token. The [LME] token is an important design detail because we use it to extract a numerical representation of the log from the neural network, further given as input to the anomaly detector. An important advantage of our method over the related work is that ADLILog does not depend on log parsing (a preprocessing procedure that extracts event templates from raw logs)~\cite{ParsingSurvey}. Since the existent log parsers are imperfect, the incorrect parsing adds additional noise and can degrade the anomaly detection performance~\cite{LogRobust}. By directly learning features from the raw logs, we eliminate this source of errors. Finally, different logs can have a variable number of tokens while the neural network requires fixed-size input. Therefore, we specify a hyperparameter $max\_len$ to unify the lengths. The shorter logs are appended with a special pad token ([PD]), while the longer ones are truncated at $max\_len$ size.

\subsection{Deep Learning Framework} The deep learning framework consists of three components: 1) embedding layer, 2) encoder network from Transformer architecture~\cite{Transformer} and 3) classification layers. Given the preprocessed and tokenized logs at the input, the embedding layer transforms the input tokens into numerical vector representations, which we refer to as vector token embeddings. The token embeddings are numerical features represented in a suitable format for the neural network. We then use the encoder network to learn relationships between the vector embeddings from the embedding layer and the appropriate target. The output from the encoder layer is the vector embedding of the input log/(static text), i.e., the [LME] vector. Depending on the training phase (pretraining or finetuning), the [LME] vector proceeds towards one of the two classification layers. The output from the classification layers is used as input in the appropriate loss function. After finetuning, the output from the second set of classification layers is the final vector embedding of the input log, which proceeds towards the anomaly detector.

\subsubsection{Embedding Layer} The \textit{embedding layer} receives the preprocessed logs as input. It serves as an interface between the textual and numerical token representation format. Specifically, each token is assigned a single index corresponding to a token embedding vector. The embeddings are learned during pretraining and are adjusted to learn the properties of the normal and abnormal events. The embeddings are learned jointly with the parameters of the neural network. Notably, the embedding layer is updated just during the pretraining phase. Updating the embedding layer during the two phases is challenged by the appearance of unseen words during finetuning. For example, there may be some operations in the target system that are not covered by the SL data, which leads to the appearance of missing tokens. The joint training with the new tokens requires an update of the internal structure of the neural network and learning new parameters every time we encounter new words. Therefore, the effective transfer of the parameters between the two learning phases is challenged. To address this issue, we introduce a special token referred to as an unknown ([UNK]) token. Notably, during pretraining (first training phase), we randomly sample 15\% of the SL data and in each sample, we replace 20\% of the tokens with [UNK] (a similar strategy is used in related works from general language~\cite{BERT}). Therefore, the pretrained model learns contexts with missing tokens. During finetuning, whenever we encounter a new token from the target-system data, we replace it with the [UNK], effectively handling the new tokens.

\subsubsection{Log Message Encoder} As a suitable architecture for the log message encoder, we identified the encoder of the Transformer~\cite{Transformer} architecture. This architecture provides state-of-the-art results in many NLP tasks (e.g., sentiment analysis, translation)~\cite{BERT}. By pointing to the similarities between the log static texts and natural language~\cite{PinjiHe2018}, we justify our design of choice. The encoder implements a multi-head self-attention mechanism that exploits the relations between tokens within the log instructions static texts. This property enables learning discriminative features between the words and the different contexts they appear in (e.g., diverse vocabularies, intent). The embedding vectors and the encoder parameters are updated via the backpropagation algorithm during pretraining. At the output of the encoder, we provide the vector embedding of the [LME] token. Due to the architectural design, the vector of the [LME] token attends over all the other token vectors during training. We considered this implementation architectural design detail because it allows learning the most relevant information from the input concerning the normal and abnormal events. The \textit{model size}, \textit{number of heads} in the encoder, and the \textit{number of encoder layers} are three hyperparameters of the log message encoder.

\subsubsection{Classification Layers} The \textit{classification layers} as input receive [LME] tokens from the encoder. It is composed of two sets of linear neural layers. As depicted in \figurename~\ref{fig:arch}, the first layer set (\textit{Set 1}) has two linear layers, with parameters $\theta^{'}$. It is trained jointly with the log message encoder during the pretraining procedure. The size of the first linear layer (from the first set of linear layers) is equal to the \textit{model size} of the encoder layer, while the second layer (from the first set of linear layers) has two neurons that correspond to the "normal" and "abnormal" severity groups from the SL data. The output of the first set of classification layers is given towards the binary cross-entropy as a loss function during pretraining. 

The second set of classification layers, with parameters $\theta^{"}$, has two linear layers (\textit{Set 2} in \figurename~\ref{fig:arch}). The two layers have the same number of neurons equal to the \textit{model size}. The output of the second set of linear layers is given as input for the loss function during finetuning. Additionally, the output of this layer is used as the final log representation, proceeded as input to the anomaly detector.

\subsubsection{Learning Process} The learning process is split into two sequential phases: pretraining and finetuning. During the \textbf{pretraining} phase, we update the parameters of the embedding layer, the log message encoder and the first set of classification layers. We perform the pretraining with the SL data, using the binary cross-entropy as a commonly used loss for binary classification~\cite{deepLearningBook2016}. The pretraining is terminated when a lack of loss improvement is observed for five consecutive epochs on a separate validation SL set. After pretraining, the parameters of the encoder and the embedding layer (as \textit{pretrained model}) are frozen and no longer updated. Thereby, they preserve anomaly-related information. The pretrained model is used to extract the initial log representation in the finetuning phase. 

For the \textbf{finetuning} phase, we pair the pretrained model with the second set of linear layers. Notably, the training data in this phase consists of the target-system data and the "abnormal" severity group from the SL data (as its subset). Since by definition of the anomaly detection task, the majority of the target-system data is assumed to describe normal system behaviour (i.e., class 0) by considering the "abnormal" class of the SL data as anomalous (i.e., class 1), the finetuning can be addressed as a binary classification problem.  The "abnormal" class of the SL data is always available, thereby, ADLILog does not need manually labeled target-system data, i.e., its unsupervised method.  Consequently, ADLILog addresses the challenge of \textbf{time-expensive} labeling. In the finetuning phase, we update just the parameters of the second set of linear layers ($\theta^{"}$), while the pretrained model is used to extract the log embeddings of the input data. The finetuning enables learning the specifics of the target-system data while relying on the anomaly-related information from the SL data. In addition, since the normal target-system data and the normal events from the SL data can differ, the finetuning adjusts the log representation embeddings to these differences. 

Another important aspect of the finetuning phase is the choice of the finetuning loss. It determines the form of the final learned log vector embeddings. Since the finetuning is defined as a binary classification problem, multiple loss choices are possible (e.g., binary cross-entropy~\cite{deepLearningBook2016}, or hyperspherical loss~\cite{hypersphericalClassification}). The binary-cross entropy is a formidable choice if the anomalous labels originate from target-system data because it allows learning of the exact discriminative properties between the classes. However, in the case of logs, the expensive labeling process makes this assumption hard. In contrast, hyperspherical loss concentrates the normal class around a single point, e.g., the centre of the hypersphere. At the same time, it is scattering the anomalous logs further apart. This is known as the \textit{concentration} property~\cite{Ruff2021}. The literature on anomaly detection~\cite{Ruff2021} suggests that preserving this property when learning representations often improves performance. Consequently, the hyperspherical loss has more desirable properties for anomaly detection, and we use it as finetuning loss. Eq.~\ref{eq:hypersphere} gives its definition for a single log $l_i$:

\begin{equation}
	\resizebox{0.9\columnwidth}{!}{%
		$L_{ad}^i =(1-y_i)||g(\mathbf{x_i};\theta, \theta^{"})||^2 - y_ilog(1-exp(-||g(\mathbf{x_i};\theta, \theta^{"})||^2))$%
	}
	\label{eq:hypersphere}
\end{equation}

where $\mathbf{x_i}$ is the log representation as output from the second classification layers set, $y_i\in\{0, 1\}$ is a label for the normal target-system data or the "abnormal" SL severity class, $\theta$ and $\theta^{"}$ are parameters of the encoder and the second set of linear layers, and $g(\mathbf{x_i};\theta, \theta^{"})$ is the function learned by the network.

\subsection{Anomaly Detector}~\label{adetector}
The goal of the anomaly detector is to highlight the anomalous target-system logs represented as log vector embeddings ($\mathbf{x_i}$). It has two components, i.e., 1) an assumed target-system normality function $\tilde{p}_{ad}^{+}$, and 2) anomaly decision rule. The normality function is an assumed model of the normal target-system logs. It is a positive function, having small values for the anomalous and large values for the normal target-system logs~\cite{Ruff2021}. The form of the function depends on the type of finetuning loss. Since the chosen hyperspherical loss learns a model that places the normal logs (class 0) close to the centre of the hypersphere, the smaller distances correspond to normal system behaviour. Following the definition of the normality function, we use the reciprocal value of the Euclidean distance between the learned log representation $\mathbf{x_i}$ and the hypersphere centre (set to the origin), given by Eq.~\ref{athreshold}. The large distances of the vector representation from the centre of the hypersphere will result in small values for the normality score (denoting anomalies) and vice versa (as seen in \figurename~\ref{fig:arch}). 

\begin{equation}
    \tilde{p}^{+}_{ad}(\mathbf{x_i}) = \frac{1}{||\mathbf{x_i}-\mathbf{c}||^2}, \quad \quad \mathbf{c}=\mathbf{0}
    \label{athreshold}
\end{equation}

Finally, to detect anomalies, we apply a decision rule on top of the normality function score values of the input logs. The decision rule involves setting a decision threshold $\tilde{a}$ over the scores, such that the logs with lower normality scores are reported as anomalous. We calculate the threshold $\tilde{a}$ on a separate validation set. The validation set is composed of the normal target-system data and the "abnormal" SL class. The threshold is set to the score that maximizes the chosen performance criteria (e.g., F$_1$ score) on the validation set.

\section{Experimental Design and Evaluation}~\label{sec5}
In this section, we present the experimental evaluation of ADLILog in comparison to three state-of-the-art and four traditional log-based anomaly detection methods. We set the focus on evaluating the detection performance, as the precise detection of anomalies is a key quality indicator for real-world deployment. The performance evaluation is made on HDFS and BGL~\cite{surveyDL} (as commonly used benchmark datasets), using three performance evaluation metrics. To estimate the ADLILog's deployment complexity, we further analyze the quality and quantity of the training data and two hyperparameters of the method and the learning procedure. These experiments evaluate the practical value of ADLILog.


\subsection{Experimental Design}
We evaluate the anomaly detection performance in two separate evaluation scenarios (1) \textit{single log line} and (2) \textit{sequential} log anomaly detection. The advantage of the single line anomaly detection resides in the potential to fast diagnose anomalies because the method directly points to the potentially anomalous log. However, the large volume of logs in short time intervals can lead to bursts of reported anomalies which in certain situations can be overwhelming. To that end, we evaluate the method's performance on event sequences as well. 

\subsubsection{Datasets} BGL and HDFS are two benchmark datasets for log-based anomaly detection that are mostly used by the research community~\cite{LogRobust, surveyDL, surveyLogAnalysis}. \tablename~\ref{tab:prop} shows the key datasets properties. To find the unique log events, we used Drain~\cite{Drain},  a state-of-the-art log parsing method. Drain's hyperparameters were set as recommended by Zhu et al.~\cite{ParsingSurvey} resulting in an output of 29 and 360 unique events for HDFS and BGL, respectively. Following He, et al.~\cite{surveyLogAnalysis} we split the dataset into 80-20\% train-test split. The first, chronologically ordered 80\% were used for training (and model tuning), while the remaining 20\% were used for performance evaluation for the two datasets accordingly.

\begin{table}[!t]
\centering
\caption{Dataset Properties}
\label{tab:prop}
\resizebox{\columnwidth}{!}{
\begin{tabular}{c|c|c|c}
\hline
Dataset & Time Span & \# Logs    & \# Anomalies \\ \hline
HDFS    & 38.7 hours    & 11,175,629 & 16,838       \\ 
BGL     & 7 months     & 4,747,963  & 348,460      \\ \hline
\end{tabular}
}
\end{table}

\textbf{HDFS} contains 11,175,629 logs generated from a map-reduce tasks on more than 200 Amazon’s EC2 nodes~\cite{PCA}. Each log has a unique identifier (block\_id) for each operation such as allocation, writing, replication and deletion. After parsing, there are 29 unique events, from which ten describe anomalous events and appear just when the block\_id is anomalous. Therefore, they are indicators of an anomaly. We used this observation to create two datasets, which we refer to as HDFS-sin and HDFS-seq. HDFS-sin is composed of time-ordered logs with a label for each event if it is anomalous or not. This data allows the evaluation of methods performance in absence of external identifiers, i.e., single log anomaly detection. HDFS-seq uses the block\_id as a natural identifier to construct sequences of events, and it is used in sequential anomaly detection evaluation. 

\textbf{BGL} contains 4,474,963 logs collected from a BlueGene/L supercomputer at Livermore Lab~\cite{SuperComputer}. BGL has two important characteristics. The first BGL characteristic is the availability of labels for individual log events given by the system administrators. We use these labels as ground truth information for single log line anomaly detection. We refer to this data as BGL-sin (with 348,460 anomalous logs). The second BGL characteristic is the absence of identifiers for task sequences. To create log sequences, similar to related work~\cite{surveyLogAnalysis}, we use a time window of size $T$. Following, He et al.~\cite{surveyLogAnalysis} we set $T=$~6h, as optimal for BGL. This resulted in a total of 828 log sequences. To obtain sequence labels, similar to related work~\cite{surveyLogAnalysis}, if there is a single anomalous log in the window, the sequence is labeled as anomalous. We refer to this data as BGL-seq, and we use it for the sequential evaluation.

\subsubsection{Competing Methods} 
We compare ADLILog (for both, the single line and the sequential evaluation) with three state-of-the-art deep learning-based methods; two supervised (LogRobust and CNN) methods and one unsupervised (DeepLog)~\cite{surveyDL}. According to the log-based anomaly detection survey by Chen et al.~\cite{surveyDL}, these three methods show the best detection performance on the two benchmark datasets. We used the public implementations of the methods available open-source in a GitHub repository~\footnote{\href{https://github.com/logpai/deep-loglizer}{https://github.com/logpai/deep-loglizer}}. The three methods were evaluated with the suggested values for their hyperparameters. Since the three methods require fixed input, similar to Chen et al.~\cite{surveyDL}, we use $window\_size$ of 10 events to create fixed-size sequences and predict the next log (i.e., the stride is one).

For the sequential evaluation, alongside the three state-of-the-art methods, we further considered four traditional log-based anomaly detection methods: two supervised, i.e., Logistic Regression (LR)~\cite{LR} and Decision Tree (DT)~\cite{DT}, and two unsupervised methods, i.e., Principle Component Analysis~\cite{PCA} and LogCluster (LC)~\cite{LogCluster}. While deep learning-based methods are directly applicable for single log line~\cite{DeepLog}, we are not aware of related work that directly applies the four traditional methods on single log lines. Therefore, we do not use them in this evaluation type. To implement the baselines we used logilizer\footnote{\href{https://github.com/logpai/loglizer}{https://github.com/logpai/loglizer}}, an open-source library for log-based anomaly detection. We set the hyperparameters of these four methods as recommended by He et al.~\cite{surveyLogAnalysis}. 

\subsubsection{Performance Evaluation Metrics}
Following related work, we use three evaluation metrics (precision, recall and F$_1$) to estimate the detection performance of the compared methods~\cite{surveyDL}. \textit{Precision} shows the fraction of the correctly reported anomalies $(Precision=\frac{\#Detected Anomalies}{\#Reported Anomalies})$. \textit{Recall} shows the correctly detected anomalies that are true anomalies $(Recall=\frac{\#Detected Anomalies}{All Anomalies})$. For anomaly detection in logs, on one side, it is important not to miss anomalies (missing an anomaly can lead to severe outages). On the other side, reporting many false positives overwhelms the operators, leading to alarm fatigue~\cite{Hassan2020}, making the method's practical usability questionable. Therefore, a natural trade-off between precision and recall emerges. In this regard, we considered $F_1$ (a harmonic mean between precision and recall $F_1=\frac{2\times Precision \times Recall}{Precision + Recall})$ as the primary evaluation metric.

\subsubsection{ADLILog Experimental Setup} We have performed the experiments using three different values for the \textit{model size} $\{16, 64, 256\}$. Using the model size of 16, we have obtained the best predictive performance. The $max\_len$ parameter was set to 32 because this length covers the majority of the log lengths. To prevent overfitting, we used the dropout regularization technique with a probability rate of $0.05$. In the pretraining phase, we used Adam~\cite{Adam} optimizer with a learning rate $10^{-4}$ and values for $\beta_1$ and $\beta_2$ set to 0.9 and 0.99. Also, we explored four different values for the batch size $\{32, 64, 256, 512\}$. The batch size of 512 showed the best average results. The finetuning was performed for five epochs with the same values for the optimizer. The experiments were conducted on a machine using Ubuntu 18.04, with CPU Intel(R) i5-9600K, RAM 128 GB, and GPU RTX 2080.

\begin{table}[!t]
\caption{Single Line Log Anomaly Detection Comparison}
\label{tab:singleleog}
\centering
\resizebox{\columnwidth}{!}{
\begin{tabular}{l|ccc|ccc}
\hline
\multicolumn{1}{c|}{\begin{tabular}[c]{@{}c@{}}Single Log Line\\ window size: 10 \\ stride: 1\end{tabular}} & \multicolumn{3}{c|}{BGL-sin}                                            & \multicolumn{3}{c}{HDFS-sin}                                           \\ \hline
\multicolumn{1}{c|}{Method}                                                          & \multicolumn{1}{c|}{F1}   & \multicolumn{1}{c|}{Prec.} & Recall & \multicolumn{1}{c|}{F1}   & \multicolumn{1}{c|}{Prec.} & Recall \\ \hline
\multicolumn{1}{l|}{ADLILog}                                                               & \multicolumn{1}{c|}{\textbf{0.61}} & \multicolumn{1}{c|}{0.55}      & 0.70   & \multicolumn{1}{c|}{\textbf{0.98}} & \multicolumn{1}{c|}{1.00}      & 0.96   \\ 
\multicolumn{1}{l|}{DeepLog}                                                           & \multicolumn{1}{c|}{0.21} & \multicolumn{1}{c|}{0.12}      & 0.82   & \multicolumn{1}{c|}{0.35} & \multicolumn{1}{c|}{0.62}      & 0.24   \\ 
\multicolumn{1}{l|}{LogRobust}                                                         & \multicolumn{1}{c|}{0.37} & \multicolumn{1}{c|}{0.63}      & 0.26   & \multicolumn{1}{c|}{0.89} & \multicolumn{1}{c|}{1.00}      & 0.79   \\ 
\multicolumn{1}{l|}{CNN}                                                               & \multicolumn{1}{c|}{0.56} & \multicolumn{1}{c|}{0.47}      & 0.68   & \multicolumn{1}{c|}{0.88} & \multicolumn{1}{c|}{1.00}      & 0.78   \\ \hline
\end{tabular}
}
\end{table}

\subsection{Experimental Results and Discussion}
The performance evaluation of the proposed approach is made using two independent experimental scenarios: single line and sequential log anomaly detection. For the \textbf{single log line anomaly detection methods comparison}, we compared ADLILog to the three state-of-the-art deep learning-based approaches. \tablename~\ref{tab:singleleog} presents the results. ADLILog significantly outperforms the supervised methods on almost all evaluation metrics. In particular, ADLILog showed the best predictive performance in terms of recall on both datasets (BGL-sin, and HDFS-sin). While DeepLog is being slightly better on recall on BGL-sin, it has significantly worsened performance on precision. On the $F_1$ as a primary evaluation metric, our method outperforms the supervised methods between 5-24\% and the unsupervised one by 40-63\%. The improvements of ADLILog are predominantly due to the rich set of abnormal events from the many diverse log instructions that help to discriminate the anomalous logs. ADLILog has a significant practical advantage in comparison to the competing supervised approaches because \textit{does not require labeled target-system anomalies}. Therefore, it can be directly applied to a target system while obtaining detection performance similar or even better than the supervised approaches (which will still require expensive manual labeling). Therefore, the good performance of ADLILog comes at a smaller practical cost.

Further, we noted that the predictive performances of all the methods for the BGL-sin dataset are significantly lower compared to the HDFS-sin. For example, the F$_1$ score of LogRobust from 0.89 on HDFS-sin falls to 0.37 on BGL-sin. A potential explanation for this observation is that the logs from BGL-sin originate from many different simultaneously running tasks. Therefore, there is a large difference between the local log neighbourhood (nearby logs) of the log subject to analysis, a phenomenon in log analysis literature known as unstable sequences~\cite{LogRobust}. In BGL-sin there are 26.94\% new log events in the test data compared to the training data, which additionally diversifies the local neighbourhoods of the events. Although some of the new events are normal, the methods that exploit the local context miss-detect them as anomalous (e.g., as seen by the drop in precision for DeepLog on BGL-sin). DeepLog, as the state-of-the-art unsupervised method, leverages the local context (window size of 10 events) to detect anomalies and it is significantly affected by the unstable sequences, resulting in the lowest performance. LogRobust and CNN leverage supervised information about the events, which helps to improve the performance. In contrast, the HDFS-sin dataset is characterized by high regularity in the sequences due to the data generation procedure. The repetitiveness of task operations (e.g., deletion, allocation), the smaller number of events and the low context change lead to higher regularity in the local contexts, which increases the detection performance. ADLILog is not affected by the local contexts differences because it examines each log independently. 

\begin{table}[!t]
\caption{Sequential Log Anomaly Detection Comparison}
\label{tab:sequenceevaluation}
\resizebox{0.5\textwidth}{!}{
\begin{tabular}{l|ccc|ccc}
\hline
\multicolumn{1}{c|}{\begin{tabular}[c]{@{}c@{}}Log Sequences\\ window size=10, \\ stride=1\end{tabular}} & \multicolumn{3}{c|}{\begin{tabular}[c]{@{}c@{}}BGL-seq\\ (time window = 6h)\end{tabular}} & \multicolumn{3}{c}{\begin{tabular}[c]{@{}c@{}}HDFS-seq\\ (block ids)\end{tabular}}       \\ \hline
\multicolumn{1}{c|}{Method}                                                                 & \multicolumn{1}{c|}{F1}              & \multicolumn{1}{c|}{Prec.}    & Recall       & \multicolumn{1}{c|}{F1}            & \multicolumn{1}{c|}{Prec.}     & Recall        \\ \hline
\multicolumn{1}{l|}{ADLILog}                                                                    & \multicolumn{1}{c|}{\textbf{0.86}}   & \multicolumn{1}{c|}{{\ul 0.84}}   & {\ul 0.88}   & \multicolumn{1}{c|}{0.93}          & \multicolumn{1}{c|}{0.92}          & 0.94          \\ 
\multicolumn{1}{l|}{DeepLog}                                                                    & \multicolumn{1}{c|}{0.63}            & \multicolumn{1}{c|}{0.46}         & 1.00         & \multicolumn{1}{c|}{0.94}          & \multicolumn{1}{c|}{0.96}          & 0.93          \\ 
\multicolumn{1}{l|}{LC}                                                                          & \multicolumn{1}{c|}{0.57}            & \multicolumn{1}{c|}{0.42}         & 0.87         & \multicolumn{1}{c|}{0.80}          & \multicolumn{1}{c|}{0.87}          & 0.74          \\ 
\multicolumn{1}{l|}{PCA}                                                                         & \multicolumn{1}{c|}{0.55}            & \multicolumn{1}{c|}{0.50}         & 0.61         & \multicolumn{1}{c|}{0.79}          & \multicolumn{1}{c|}{0.98}          & 0.67          \\ 
\multicolumn{1}{l|}{LogRobust}                                                                    & \multicolumn{1}{c|}{0.83}            & \multicolumn{1}{c|}{0.71}         & 1.00         & \multicolumn{1}{c|}{0.96}          & \multicolumn{1}{c|}{0.93}          & 0.98          \\ 
\multicolumn{1}{l|}{CNN}                                                                     & \multicolumn{1}{c|}{0.82}            & \multicolumn{1}{c|}{0.69}         & 1.00         & \multicolumn{1}{c|}{0.97}          & \multicolumn{1}{c|}{0.94}          & 0.99          \\ 
\multicolumn{1}{l|}{LR}                                                                           & \multicolumn{1}{c|}{0.71}            & \multicolumn{1}{c|}{0.95}         & 0.57         & \multicolumn{1}{c|}{0.98}          & \multicolumn{1}{c|}{0.95}          & 0.99          \\ 
\multicolumn{1}{l|}{DT}                                                                           & \multicolumn{1}{c|}{0.72}            & \multicolumn{1}{c|}{0.95}         & 0.57         & \multicolumn{1}{c|}{\textbf{1.00}} & \multicolumn{1}{c|}{\textbf{1.00}} & \textbf{1.00} \\ \hline
\end{tabular}
}
\end{table}

ADLILog detects anomalies into single log lines. To be able to detect anomalies in \textbf{log sequences}, ADLILog aggregates the predictions of the individual logs from the sequence. If there is at least a single detected anomalous log, the whole sequence is detected as anomalous. For BGL-seq, we aggregated ADLILog's logline predictions into the fixed time intervals (same to BGL-seq data generation procedure), while for HDFS-seq, we aggregate the predictions based on the block\_ids. This way, ADLILog additionally can be evaluated on sequential log anomaly detection. 

\tablename~\ref{tab:sequenceevaluation} gives the results of the \textbf{sequential log anomaly evaluation}. We discuss them for each dataset individually, starting with BGL-seq. For BGL-seq, ADLILog achieves the best performance among the deep-learning state-of-the-art and the traditional approaches on the $F_1$ evaluation metric. Considering precision, ADLILog outperforms the three unsupervised (DeepLog, LC and PCA) and the two deep-learning supervised methods, while being outperformed by DT and LR. In contrast, when considering recall, ADLILog outperforms all traditional approaches. Since the logs from the BGL-seq dataset resemble a real-world behaviour of a supercomputer, the logs originate from many different independent tasks and are intertwined. This leads to low repetitiveness of the same log sequences. Therefore, the state-of-the-art methods (LogRobust, DeepLog and CNN), which directly model the sequences face the challenge of unstable sequences. Line of works~\cite{surveyDL, LogRobust} also shows that the unstable sequences significantly affects the performance of log anomaly detection methods. In contrast, ADLILog focuses on the discriminative properties of the individual events, ignoring the sequential features of the anomalous patterns. These experimental results show that in the case of lower repetitiveness in the log sequences, leveraging solely the differences in the language used to describe normal and anomalous events can lead to better anomaly detection performance. 

The results on HDFS-seq show that the methods exploiting sequential properties achieve better results. The key characteristic of HDFS-seq is the high sequence regularity which explains the good performance of the sequential methods. In addition, not all anomalies are logged into a single log. For example, around 70\% of the anomalies in the test dataset can be identified by shortened log sequences, however, the majority of them have at least one anomalous logline. Since ADLILog does not model the sequential properties directly, it does not learn the anomalous sequence properties (e.g., long sequences caused by delays where no anomalous events are recorded). Nevertheless, it outperforms the traditional unsupervised baseline methods by simply relying on the language properties of the logs. The supervised methods are showing stronger performance in comparison to the unsupervised. However, in comparison to ADLILog (which does not require labeled data), the supervised methods require expensive labeling. ADLILog has strong anomaly detection performance for the two log datasets with weak and strong sequential dependencies. Notably, a significant practical improvement of ADLILog over the related methods is that it does not require labels from the target system to learn a model. Therefore, ADLILog achieves competitive detection performance at a lower practical cost.


\begin{figure}[!h]
\centering
\includegraphics[width=0.7\columnwidth]{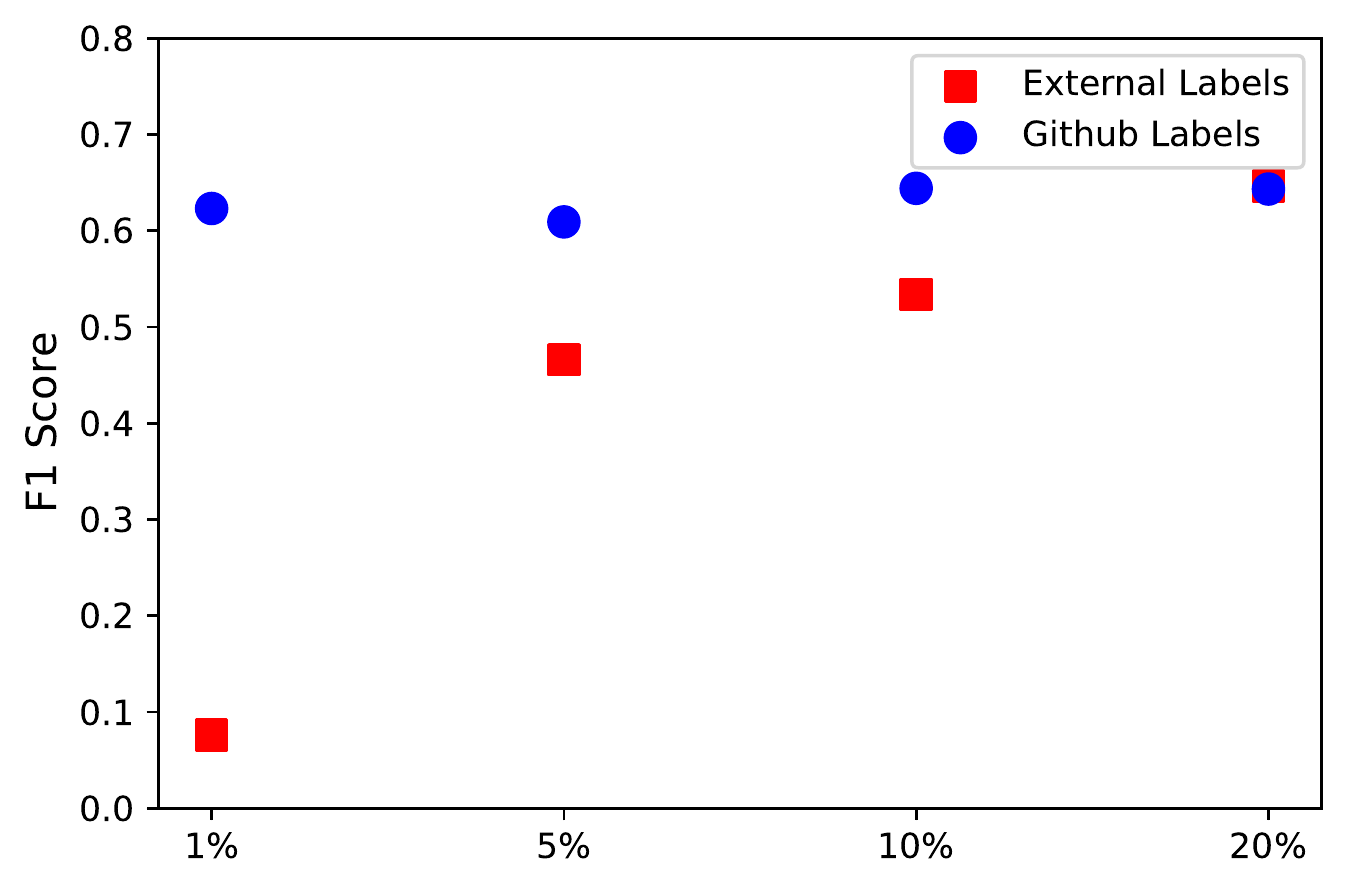}
\caption{Quantitative and qualitative evaluation of the "Abnormal" logs}
\label{fig:results:qualitativelabels}
\end{figure}

The acceptance of AI-enabled methods in production settings depends strongly on the necessary effort for creating training data. Therefore, we \textbf{examine the quantitative and qualitative properties of the training data} ADLILog needs to learn a good model. Specifically for the quantitative property, we varied the ratio of normal versus anomalous logs in the training data when finetuning. This experiment examines the relative ratio between normal target-system and "abnormal" logs needed for finetuning. For the qualitative property, we varied the origin of the anomalous class (class 1), i.e., if it comes from human-provided labels from other software systems (external system labels) or the "abnormal" class of the SL data (GitHub Labels). The external system labels are obtained from publicly available datasets (e.g., TBIRD and SPIRIT~\cite{SuperComputer} -- two supercomputer datasets with labeled anomalies). Similarly, as the adoption of the "abnormal" class in finetuning, the external-system labels denote anomalous concepts from related systems (thus eliminating the requirement for labeling), and can be used for model finetuning (in place of the "abnormal" GitHub class). \figurename~\ref{fig:results:qualitativelabels} depicts the experimental results for quantitative evaluation when incrementally changing the ratio between the anomalous and normal logs in the training data for BGL-sin in the ranges $\{1\%, 5\%, 10\%, 20\%\}$. They show that with a small ratio of 1~\%, ADLILog achieves $>95\%$ of the optimal performance on the $F_1$ metric. The improvement is consistent with further increasing the ratio. Concerning the qualitative evaluation, by varying the two sets of labels, the results show that the GitHub labels provide robust and better performance because of the greater semantic variety in the "abnormal" events over the external system labels. This experiment demonstrates that a small number of labels from the "abnormal" class of the SL data have significant practical usage for log anomaly detection. 

\begin{figure}[!h]
\centering
\includegraphics[width=0.8\columnwidth]{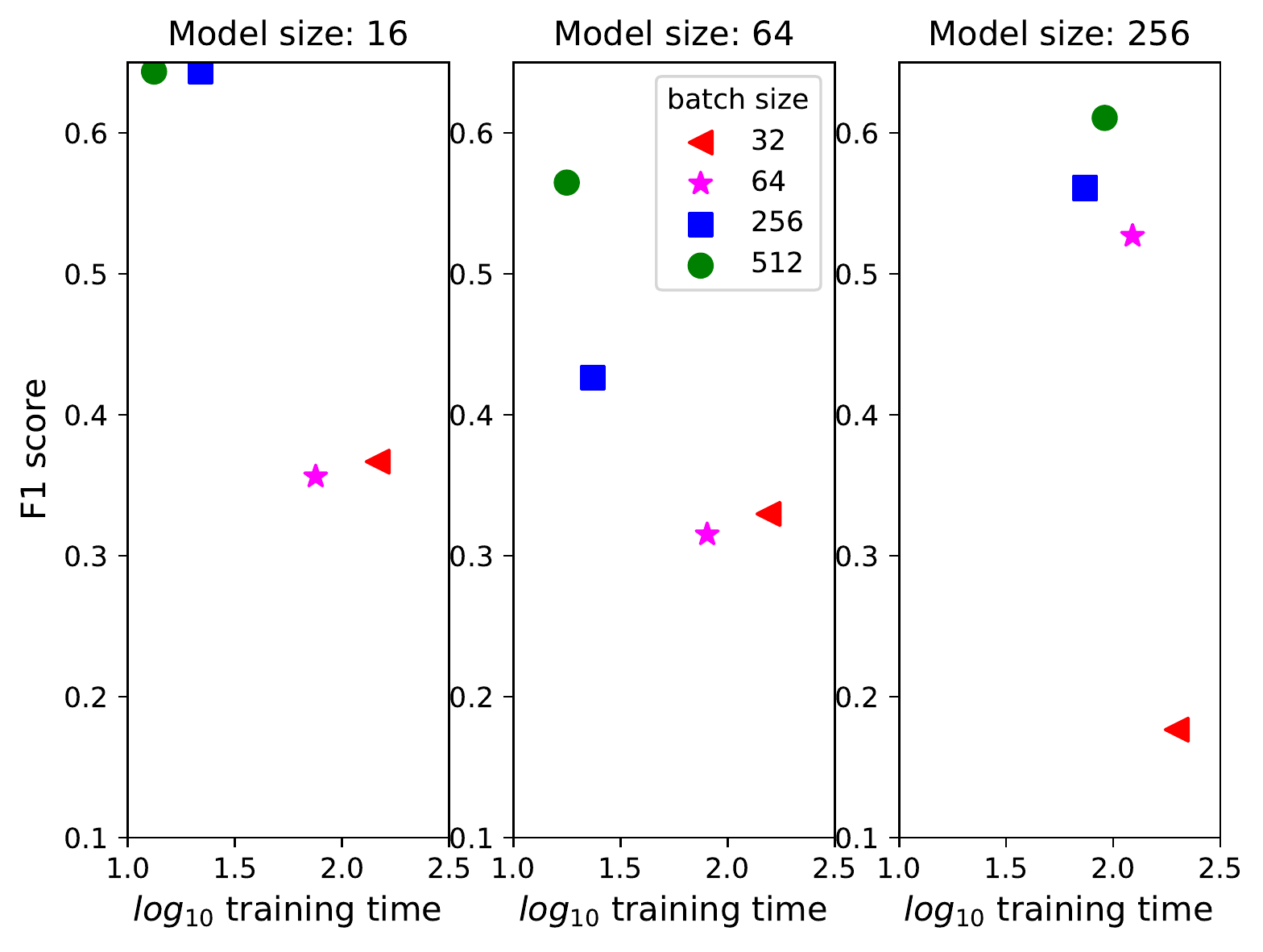}
\caption{Sensitivity analysis of the influence of batch and model size}
\label{fig:results:parameterstudy}
\end{figure}

Finally, the correct parameter setting influences the needed effort for fast configuration and the quality of the detection. To evaluate the \textbf{impact of the hyperparameters over the detection performance and efficiency}, we examined two hyperparameters, model and batch sizes, influencing the model performance and update time. We considered the BGL-sin dataset. The experimental results when varying the model size in the range $\{16, 64, 256\}$ and batch size in the range $\{32, 64, 256, 512\}$, reported in \figurename~\ref{fig:results:parameterstudy}, show that the larger batch size and smaller model size provide better detection performances while being faster for updating. The prediction time per batch size of 512 is 17~ms ($\sim$30000 logs per second). Together with the small model size, these experiments imply that ADLILog has desirable practical properties.

\section{Related Work}\label{rw}
There exist multiple methods for the automation of log-based anomaly detection~\cite{LR, LogRobust, DT, DeepLog}. From the perspective of target-system log labels availability, the methods are categorized into supervised and unsupervised. The supervised methods assume the existence of labels from the target system when learning a model. In one of the earliest applications of these methods, Bodik et al.~\cite{LR} applied Logistic Regression (LR) to successfully detect anomalies in data centres, by treating the problem as a binary classification. Decision Trees (DT)~\cite{DT} were utilized in the detection of anomalous web requests from access logs. These two methods start by log parsing to extract events and then use count vectors in a fixed time interval as input samples. The recent advances in deep learning resulted in the appearance of several supervised deep-learning-based methods, e.g., LogRobust~\cite{LogRobust}, and CNN~\cite{CNN}. LogRobust uses the LSTM architecture, augmented with attention. These two are popular deep learning architectures frequently combined for sequence modeling. LogRobust, as input, receives a sequence of events, and as output, it predicts if the observed sequence is anomalous or not. By careful sequence construction, i.e., by incremental sliding over the log sequences by one element, it can be used to predict single log lines~\cite{DeepLog}. An additional feature of LogRobust is using vector embeddings from general-purpose languages to represent the logs. Lu et al.~\cite{CNN} use Convolutions Neural Networks (CNN), another type of deep learning architecture, to learn normal and abnormal sequences from template indices. Similar to LogRobust, CNN can be used to detect anomalies from a single log. While having strong detection performance, the large frequency of the software updates and the large volume of the produced logs make the labeling process expensive. Therefore, supervised methods are frequently considered impractical~\cite{surveyLogAnalysis}.

In contrast, the unsupervised methods do not assume the existence of labeled data. This has an important practical implication because it eliminates the need for expensive labeling. Therefore, the unsupervised anomaly detection methods are easier to adopt.  In one of the earliest works on unsupervised anomaly detection, Xu et al. apply PCA~\cite{PCA} to learn the normal state of the event counts by projecting them as points in a vector space. In the test phase, the test sample is projected in the constructed vector space and reported as an anomaly if the projection significantly deviates from the learned normal state. Lin et al.~\cite{LogCluster} introduce LogCluster which uses the TFIDF algorithm for sequence representation. It first constructs a knowledge base of normal/anomalous sequence clusters by agglomerative clustering and human-based cluster labeling (normal or anomalous). A test sample is detected as an anomaly if it is clustered into anomalous clusters. DeepLog~\cite{DeepLog} and LogAnomaly~\cite{LogAnomaly} are two popular unsupervised deep learning-based methods. The innovative feature of the two is the introduction of an auxiliary task called "next event prediction" (NEP). NEP is a supervised task that given a sequence of events, forecasts the most probable next event. Notably, the labels originate from the input data itself, i.e., no labeling is performed, which makes the methods unsupervised. The test sequences with an incorrect prediction for the next event are considered anomalous. As stated by the authors, DeepLog can be applied for sequential and single logs given as inputs. To learn the normal state, an LSTM architecture is trained on the NEP task. LogAnomaly has two additional features: 1) it uses log semantics and 2) event counts in joint training a DeepLog-like model. The empirical results show that the improvements over DeepLog are not significant~\cite{surveyDL, LogAnomaly}. Unsupervised methods are often criticized for their lower performance in comparison to the supervised ones~\cite{surveyLogAnalysis} leading to alarm fatigue and discouraging their wide applicability.

In addition, there are other methods for log-based anomaly detection in both industry and research~\cite{Logsy, SwissLog, HuaweiFLAP}. However, those solutions are part of production systems~\cite{HuaweiFLAP} or have specific implementation challenges while do not provide public implementations~\cite{SwissLog, UniLog2021, Logsy}. Due to the inability of transparent comparison, we do not discuss them in detail. 

\section{Conclusion}~\label{sec6}
This paper addresses the problem of automating log-based anomaly detection as a crucial maintenance task in enhancing the reliability of IT systems. It introduces a novel unsupervised method for log anomaly detection, named ADLILog. The key idea of ADLILog is to use the large unstructured information from the logging instructions of 1000+ GitHub public code projects to improve the target-system log representations, which directly improves anomaly detection. We first conducted a study to examine the language properties of the log instructions, and we show that they encode rich anomaly-related information. ADLILog combines the anomaly-related information and the target-system data to learn a deep neural network model by a sequential two-phase learning procedure. The extensive experimental results on the two most commonly used benchmark datasets show that ADLILog outperforms the related methods: the supervised by 5-24\%, and the unsupervised by 40-63\% on the F$_1$ score. Further experiments demonstrate that ADLILog has desirable practical properties concerning the time-efficient model updates and small model sizes. This study signifies the benefit of using large unstructured information in aiding the automation of IT operations. Regarding future work, the paper opens additional questions on how to apply the SL data for automating higher-order IT operational tasks, like failure identification and root-cause analysis.
\bibliographystyle{IEEEtran}
\bibliography{IEEEabrv,main.bib}
\end{document}